\documentclass[10pt,letterpaper]{article}

\usepackage{cogsci}
\usepackage{booktabs}
\usepackage{tabularx}
\usepackage{graphicx}
\usepackage{authblk}

\usepackage{amssymb}

\cogscifinalcopy

\usepackage{multirow}
\usepackage[
  style=apa,
  natbib=true,
  annotation=false,
]{biblatex}
\usepackage{float} 

\title{\textbf{CogEvolution: A Human-like Generative Educational Agent to Simulate Student's Cognitive Evolution}}

\author[1,2]{\mbox{Wei Zhang}}
\author[1,2]{\mbox{Yihang Cheng\textsuperscript{*}}}
\author[1,3]{\mbox{Zhirong Ye}}
\author[1,3]{\mbox{Kezhen Huang}}
\affil[1]{Faculty of Artificial Intelligence in Education, Central China Normal University, Wuhan, Hubei, China}
\affil[2]{National Engineering Research Center for Educational Big Data, Central China Normal University, Wuhan, Hubei, China}
\affil[3]{Central China Normal University Wollongong Joint Institute, Central China Normal University, Wuhan, Hubei, China}
\affil[*]{Corresponding author:yihangcheng@mails.ccnu.edu.cn}

\begin{document}

\maketitle
\begin{abstract}
Generative Agents, owing to their precise modeling and simulation capabilities of human behavior, have become a pivotal tool in the field of Artificial Intelligence in Education (AIEd) for uncovering complex cognitive processes of learners. However, existing educational agents predominantly rely on static personas to simulate student learning behaviors, neglecting the decisive role of deep cognitive capabilities in learning outcomes during practice interactions. Furthermore, they struggle to characterize the dynamic fluidity of knowledge internalization, transfer, and cognitive state transitions. To overcome this bottleneck, this paper proposes a human-like educational agent capable of simulating student cognitive evolution—\textbf{CogEvolution}. Specifically, we first construct a cognitive depth perceptron based on the Interactive, Constructive, Active, Passive (ICAP) taxonomy from cognitive psychology, achieving precise quantification of learner cognitive engagement. Subsequently, we propose a memory retrieval method based on Item Response Theory (IRT) to simulate the connection and assimilation of new and prior knowledge. Finally, we design a dynamic cognitive update mechanism based on evolutionary algorithms to simulate the real-time integration of student learning behaviors and cognitive evolution processes. Comprehensive evaluations demonstrate that CogEvolution not only significantly outperforms baseline models in behavioral fidelity and learning curve fitting but also uniquely reproduces plausible and robust cognitive evolutionary paths consistent with educational psychology expectations, providing a novel paradigm for constructing highly interpretable educational agents.

\end{abstract}

\section{Introduction}

In recent years, breakthrough advancements in Large Language Models (LLMs) have provided novel opportunities for the deep integration of cognitive science and educational technology \citep{ luo2025largelanguagemodelagent,dai2025psycher1}. As a significant application form of LLMs, Generative Agents demonstrate immense potential in simulating complex human behaviors due to their superior capabilities in memory retrieval, planning, reasoning, and reflection \citep{park2023generative, xi2025rise, sumers2024cognitive}. In cognitive science research, these agents are widely used to construct computational social models to verify sociological hypotheses or reproduce human decision-making processes in specific contexts \citep{aher2023using, chuang2024wisdom}.

With deepening research, the application of generative agents in educational scenarios has become increasingly prominent, particularly in simulating dynamic interactions within teaching relationships. Researchers have begun dedicating efforts to utilizing such agents to build high-fidelity "Student Simulations" \citep{markel2023gpteach, xu2024eduagent}. By learning from real student response logs and interaction histories, these agents can assist in evaluating the effectiveness of instructional strategies or serve as standardized patients for training novice teachers \citep{xu2024eduagent}, thereby enhancing educational quality in low-risk environments.

However, despite progress in simulating explicit behaviors, agents still face severe challenges in modeling deep cognitive mechanisms. Frontier research on EvoAgents \citep{li2025evoagents} has acutely pointed out the issue of \textbf{"Cognitive Dynamics Disconnect"} present in generative agents. Specifically, agents lacking theoretical support from cognitive psychology struggle to possess the capability to truly simulate human internal mental activities. For example, mainstream architectures such as SmallVille \citep{park2023generative} neglect the cognitive architectures driving these behaviors \citep{sumers2024cognitive}.

In educational scenarios, this "cognitive disconnect" phenomenon is particularly pronounced, which we specifically define as \textbf{"Missing Cognitive Features"} and \textbf{"Cognitive Path Stagnation."} Current educational agent architectures largely rely on \textbf{Static Personas} constructed based on response behavior features. Whether they are base agents constructed directly using native large models like Llama 3 \citep{meta2024llama3}, or behavioral modeling by LLM combined with deep learning neural networks \citep{lv2025genal}, their essence lies in the parametric fitting of a snapshot of student ability. These static parametric methods neglect the essence of learning—a dynamic process of continuously reconstructing cognitive schemas through Assimilation and Accommodation \citep{piaget1976piaget}. They struggle to capture the dynamic fluidity of knowledge internalization, ability transfer, and cognitive states (such as from confusion to insight) generated by deep cognitive engagement (such as constructive and interactive modes in ICAP theory \citep{chi2014icap}) during practice. To bridge this gap, this paper proposes the \textbf{CogEvolution}. This is the first computational agent based on educational psychology theories that focuses on the dynamic evolution of \textbf{cognitive states} and \textbf{cognitive paths} in agent learning activities. 

Specifically, the main contributions of this paper are summarized as follows:
\begin{enumerate}
    \item Refined the definition of the cognitive dynamics disconnect problem in the educational domain and proposed the CogEvolution, pioneering a research paradigm for exploring cognitive evolutionary paths in generative agents.
    \item Innovatively integrated the ICAP cognitive perception module as the adapter of CogEvolution, designed a memory retrieval based on IRT theory\citep{lord1980applications}, and a state update mechanism inspired by evolutionary algorithms, realizing a paradigm shift from "static persona" to "dynamic cognitive flow."
    \item Constructed the CogMath-948 dataset containing cognitive state annotations and demonstrated through comprehensive evaluation that CogEvolution significantly outperforms existing baseline models in learning curve fitting and knowledge transfer capabilities.
\end{enumerate}

\section{Related Work}

With Large Language Models (LLMs) demonstrating powerful reasoning and generalization capabilities, research on Generative Agents has experienced explosive growth \citep{xi2025rise, luo2025largelanguagemodelagent,dai2026tearsorcheers}. The "Stanford Town" simulation proposed by Park et al. \citep{park2023generative} is a milestone work in this field, constructing an architecture containing Memory Stream, Reflection, and Planning, enabling agents to simulate long-term human social interactions. Subsequently, this paradigm was widely extended: Wang et al. \citep{wang2024survey} systematically reviewed the potential of LLM-based autonomous agents in task planning and tool use; Li et al. \citep{li2025evoagents} proposed EvoAgents, which further introduced psychodynamic mechanisms, resolving the issue of rigid personalities in agent social simulations. Additionally, the emergence of open-source evaluation platforms like AgentSims \citep{sumers2024cognitive} has lowered the threshold for constructing high-fidelity social simulation environments. However, these general architectures mostly focus on the breadth of social behaviors and the coherence of interpersonal interactions, lacking fine-grained modeling of individual internal cognitive processes (such as knowledge construction and conceptual change), making them difficult to directly transfer to educational learning scenarios that require high cognitive depth.

In the field of education, utilizing agents to simulate students (Student Simulation) to assist in instructional evaluation and teacher training has become a core topic. The GPTeach platform developed by Markel et al. \citep{markel2023gpteach} uses LLMs to generate virtual students with different knowledge backgrounds for training teaching assistants' instructional response capabilities. To enhance simulation realism, Xu et al. \citep{xu2024eduagent} proposed EduAgent, which achieves multimodal cloning of student classroom behaviors (such as raising hands, disruptive behavior) by integrating audiovisual and text data from real classrooms. Although these works have achieved success in simulating explicit instructional behaviors, their core mostly relies on Static Personas or behavioral cloning \citep{mandyam2025tutortest}, meaning agents tend to generate fixed response patterns based on preset labels of "low-performing" or "high-performing" students, rather than dynamically evolving their knowledge states based on Cognitive Engagement during interactions, leading to the phenomenon of "Missing Cognitive Features."

In summary, although generative agents have made significant progress in behavioral simulation and instructional assistance, existing research still has gaps in simulating the dynamics of learner cognitive development and psychological interpretability. There is an urgent need to establish a computational method capable of real-time alignment between external learning behaviors and internal cognitive state evolution.

\section{Model Design}

\begin{figure*}[ht]
    \centering
    \includegraphics[width=\linewidth]{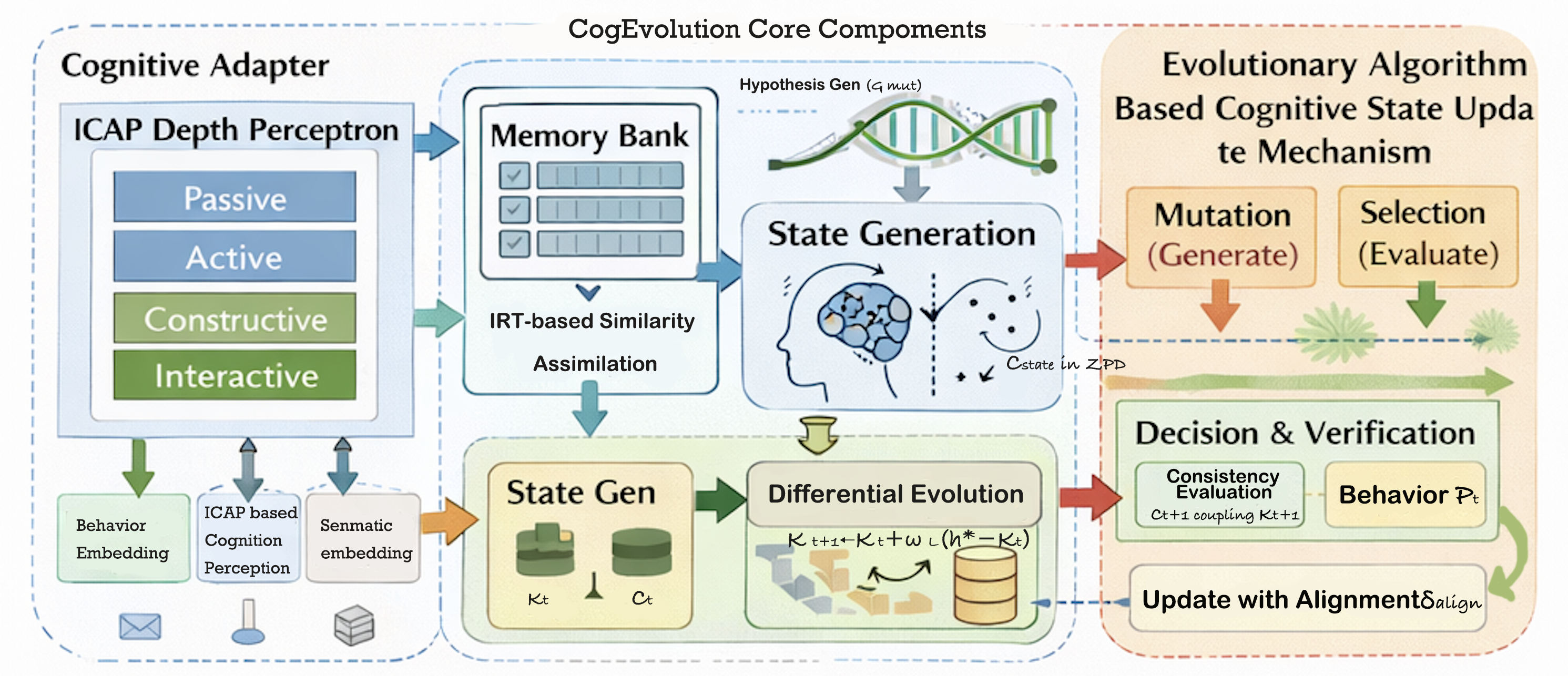} 
    \caption{Overview of the CogEvolution framework.}
    \label{fig:framework}
\end{figure*}

The CogEvolution agent aims to simulate a cognitive subject capable of dynamic evolution. As shown in Figure \ref{fig:framework}, its core logic lies in perceiving cognitive depth through the ICAP mechanism and utilizing evolutionary algorithms to drive the restructuring of knowledge structures. The Cogevolution consists of three highly coupled components: Cogitive Adapter for Depth Perceptron, an agent input adapter responsible for environmental perception and cognitive feature initialization; IRT-Driven Memory Retrieval Module, executing memory retrieval, state generation, and knowledge evolution; Evolutionary Algorithm-Based Cognitive State Update Mechanism, used to couple the simulation of learning behaviors with their corresponding cognitive paths and dynamically update cognitive evolution.

\subsection{Symbol Definition}
To provide a rigorous formal description of the cognitive evolution process, we define the symbol system as shown in Table \ref{tab:symbols}.

\begin{table}[ht]
    \centering
    \caption{Symbol Definitions and Parameter Descriptions}
    \label{tab:symbols}
    \resizebox{\columnwidth}{!}{
    \begin{tabular}{l|p{9cm}} 
        \toprule
        \textbf{Symbol} & \textbf{Description and Physical Meaning} \\
        \midrule
        $\mathcal{A}$ & Agent object \\
        $I_t$ & Input vector at time $t$ (question, logs, environmental features) \\
        $S_t$ & Agent comprehensive state, $S_t = \{P, C_t, K_t\}$ \\
        $K_t$ & \textbf{Knowledge Structure}, long-term stable cognitive schema \\
        $C_t$ & \textbf{Cognitive State}, transient state (e.g., confusion, flow) \\
        $\hat{p}_t$ & Mastery confidence probability, used for AUC/RMSE calculation \\
        $L_t$ & Cognitive Engagement Depth, $L \in \{Passive, Active, Constructive, Interactive\}$ \\
        $\mathbf{\hat{y}}_{icap}$ & Probability distribution vector over ICAP levels \\
        $v_i$ & Baseline gain coefficient for the $i$-th ICAP level \\
        $\omega_{L}$ & \textbf{Cognitive Evolution Rate}, weighted continuous ICAP intensity \\
        $\mathcal{M}$ & Memory bank storing quadruples $(I, Response, C, L)$ \\
        $\text{Sim}_{struct}$ & Structural similarity based on IRT characteristic curve distance \\
        $\text{Sim}_{sem}$ & Semantic similarity based on text embedding \\
        $\theta_t$ & Learner's latent ability parameter in IRT \\
        $P_i(\theta)$ & Item Characteristic Function (ICC) for question $i$ \\
        $\mathcal{G}_{evo}$ & Mutation Operator: Hypothesis Generation Function \\
        $\mathcal{H}$ & Set of generated cognitive hypotheses (offspring) \\
        $\sigma$ & Mutation intensity parameter modulated by emotion \\
        $\text{ZPD}$ & \textbf{Zone of Proximal Development}, constrains effective search space \\
        $\mathcal{S}_{eval}$ & Selection Operator: Consistency Evaluation Function \\
        $\gamma$ & Penalty coefficient for deviating from ZPD \\
        $\delta_{align}$ & Behavior-Cognition Alignment Coefficient \\
        \bottomrule
    \end{tabular}
    }
\end{table}

\subsection{Cognitive Adapter}
The adapter serves as the perception entry point of the agent, designed to establish the foundational information field required for simulation operation. We constructed a multimodal feature extractor to extract feature vectors $I_t$ from interaction logs from the CogMath-948 dataset into the embedding of Agent's Adapter, including:
\begin{enumerate}
    \item Cognitive Semantic Features: Utilizing Sentence-BERT to extract cognitive features and the textual information of the interactive practice in the dataset, such as incorrect classifications and reflections.
    \item Behavioral Operation Features: Including interaction types (reading, questioning, self-explanation) and response latency.
\end{enumerate}

The perceptron $\Phi_{ICAP}$ maps input features to a probability distribution across the four ICAP levels. The specific mathematical implementation is an attention-based hierarchical probability mapping:
\begin{equation}
    \mathbf{\hat{y}}_{icap} = \text{Softmax}(\mathbf{W}_c \cdot \text{Encoder}(I_t) + \mathbf{b}_c)
\end{equation}
Based on the probability distribution, we calculate the weighted continuous \textbf{Cognitive Evolution Rate} $\omega_L$ to achieve smooth state transitions:
\begin{equation}
    \omega_L = \text{SoftPlus} \left( \sum_{i \in \{P,A,C,I\}} v_i \cdot \hat{y}_i \right)
\end{equation}
where $v_i$ represents the baseline gain coefficient for each level (e.g., Constructive corresponds to 1.5). $\omega_L$ serves as a key parameter determining the step size of subsequent cognitive state updates, thereby achieving precise quantification of learner cognitive investment.

\subsection{Memory Retrieval}
This module aims to simulate the psychological mechanisms of "connecting new and old knowledge" and "Assimilation" in human learning. We introduce Item Response Theory (IRT) to define deep structural similarity.

The system calculates a hybrid similarity between the current question $q$ and memory $m \in \mathcal{M}$:
\begin{equation}
    Score(m) = \alpha \cdot \text{Sim}_{sem}(q, m) + \beta \cdot \text{Sim}_{struct}(q, m)
\end{equation}
where $\text{Sim}_{struct}$ measures the isomorphism of questions in the cognitive ability space. We define it as the integral distance between the IRT Characteristic Curves (ICC) of two questions near the learner's current ability $\theta_t$:
\begin{equation}
    \text{Sim}_{struct}(e_i, e_j) \propto \exp \left( - \int_{\theta_{t}-\delta}^{\theta_{t}+\delta} |P_i(\theta) - P_j(\theta)| d\theta \right)
\end{equation}
This formula ensures that two questions are judged as structurally similar only when they require similar levels of cognitive ability to solve.
If a memory with high structural similarity is retrieved ($Score > \tau$), the agent directly invokes the old schema to solve the new problem, at which point the cognitive state remains stable. If structural similarity is low but the perceived ICAP engagement $\omega_L$ is high, the system determines that prior knowledge is insufficient to explain the new situation, generating a cognitive state $C_t$ of "confusion" or "exploration," triggering subsequent evolutionary updates.

\subsection{Evolutionary Cognitive State Update}
This is the core engine of CogEvolution, designed to simulate the cognitive evolution process and achieve dynamic updates. We draw upon the "Mutation-Selection-Update" paradigm of evolutionary algorithms to couple learning behaviors with cognitive evolutionary paths in real-time.

\subsubsection{Mutation Operator: Cognitive Hypothesis Generation (Mutation)}
The current knowledge structure $K_t$ is treated as the parent individual. When cognitive conflict is perceived (input $I_t$ cannot be fully explained by $K_t$), the system triggers a mutation operation. Utilizing the generative capability of LLMs as the \textbf{Mutation Operator $\mathcal{G}_{mut}$}, $\lambda$ offspring hypotheses are generated based on $K_t$:
\begin{equation}
    \mathcal{H} = \{h_1, h_2, ..., h_\lambda\} = \mathcal{G}_{mut}(K_t, I_t, \sigma)
\end{equation}
where $\sigma$ is the mutation intensity, modulated by the current emotional state. These hypotheses represent the multiple possibilities generated by the learner when confused (e.g., "Is the formula wrong?" or "Is this a new question type?"), constituting a temporary population at the thought level.

\subsubsection{Fitness Evaluation \& Selection}
We define a fitness function $\mathcal{F}(h)$ to evaluate the quality of each hypothesis. To ensure the evolutionary process conforms to human cognitive laws, we introduce Vygotsky's \textbf{Zone of Proximal Development (ZPD)} theory as an evolutionary constraint\citep{vygotsky1978mind}. In the algorithm, ZPD is formalized as an effective search radius centered on the current knowledge structure $K_t$. The fitness function evaluates whether hypothesis $h$ falls within this region:
\begin{equation}
    \mathcal{F}(h) = \text{Consistency}(h, I_t) - \gamma \cdot \text{Distance}(h, K_t, \text{ZPD})
\end{equation}
Here, the first term measures logical self-consistency, and the second term penalizes "excessive leaps" or "stagnation" deviating from the ZPD. The system executes tournament selection to screen the hypothesis $h^*$ with the highest fitness:
\begin{equation}
    h^* = \operatorname*{argmax}_{h \in \mathcal{H}} \mathcal{F}(h)
\end{equation}

\subsubsection{Evolutionary Update \& Inheritance}
The final step is to generate the next-generation knowledge structure $K_{t+1}$. We adopt a Differential Evolution approach to simulate continuous evolution on the local population, integrating the features of the winning hypothesis $h^*$ into the parent with the ICAP evolution rate $\omega_L$ as the step size:
\begin{equation}
    K_{t+1} \leftarrow K_t + \omega_L \cdot (h^* - K_t)
\end{equation}
This formula indicates that the magnitude of knowledge evolution depends on the depth of cognitive engagement ($\omega_L$): passive participation leads to minor mutations, while constructive participation allows for significant structural restructuring.

\subsection{Decision and Verification}

The agent generates final external behaviors based on the updated knowledge structure $K_{t+1}$. To support the precise quantification of learning states, this module outputs two types of information:

\subsubsection{Behavioral Decision:}
Generates specific problem-solving steps or natural language responses $D_{out}$.

\subsubsection{Mastery Confidence:}
To simulate the student's anticipation of their own ability and support subsequent Knowledge Tracing assessments (such as AUC/RMSE calculation), the model outputs the predicted probability $\hat{p}_t$ that the current response is correct. We obtain this value utilizing LLM Logits or Self-Reflection Prompts:
\begin{equation}
    \hat{p}_t = P(\text{Correct} | K_{t+1}, I_t) \in [0, 1]
\end{equation}

\subsubsection{Alignment Check:}
The system calculates the semantic alignment coefficient $\delta_{align}$ between cognitive state $C_{t+1}$ and behavior $D_{out}$. If $\delta_{align}$ is below the threshold $\tau$, indicating "cognitive dissonance" (e.g., lacking confidence but forcing an answer), the system marks this interaction step as low confidence:
\begin{equation}
    \delta_{align} = \text{CosSim}(\text{Embed}(D_{out}), \text{Embed}(C_{t+1}))
\end{equation}

\section{Experimental Set Up}

To systematically verify the effectiveness of CogEvolution in simulating real learner cognitive dynamics, we designed a series of comparative experiments aimed at answering three core research questions:

\begin{itemize}
    \item \textbf{RQ1} Does CogEvolution outperform traditional static persona and knowledge tracing models in terms of ability prediction accuracy and mistake fidelity?
    
    \item \textbf{RQ2} Can CogEvolution reproduce the non-linear learning trajectories characteristic of human learners, specifically regarding conformity to the Power Law of Practice?
    
    \item \textbf{RQ3} What is the validity of the simulated cognitive evolution process, and how does each core module contribute to the overall performance?
\end{itemize}

\subsection{Dataset and Baseline Models}
Given the lack of suitable open datasets in the current research field, we constructed the \textbf{CogMath-948} dataset collected from real educational scenarios to facilitate in-depth research on fine-grained cognitive evolution. This dataset covers real learning data from 1,245 eighth-grade students over six months. For each student interaction, we collected rich information across the following four dimensions:
Question and Response Information: Includes question stem, student-selected answer, and standard correct answer;
Reflective Explanation: Textual explanations provided by students after answering, elucidating their reasoning process or causes of error, which is a key basis for inferring cognitive states.
Cognitive Engagement Label: Cognitive depth labels derived based on the ICAP taxonomy;
Fine-grained Misconception Classification: Specific error concept categories associated with incorrect answers.

\subsection{Baseline Models and Parameter Settings}
To comprehensively evaluate model performance, we selected three types of representative baseline models. 

\subsubsection{General LLMs} including Llama 3 (70B) \citep{meta2024llama3} and GPT-4o\citep{openai2025gpt4o}, representing high-performance base capabilities using standard Persona Prompting.

\subsubsection{Reasoning-Enhanced Models} such as Gemini 3 Pro Think\citep{google2025gemini3pro}, used to investigate the impact of long Chain-of-Thought on the simulation process. 

\subsubsection{Knowledge Tracing (KT)-augmented Agent} (Discriminative Assisted Generative Baseline\citep{arana-etal-2025-foundations}), we reproduced the PEERS framework as a strong baseline. This model integrates classic Bayesian Knowledge Tracing (BKT)\citep{corbett1995knowledge} with LLMs, utilizing external data-driven methods to predict student mastery probabilities and generate responses accordingly. 

Using this as a comparison group aims to demonstrate the differences in simulation authenticity between "external data-driven (KT)" and the "internal cognitive evolution (CogEvolution)" proposed in this paper.

\subsection{Metrics}
This study adopts a two-stage evaluation system, covering ability simulation accuracy and cognitive evolutionary scientificity:

\subsubsection{Task Performance:}
\begin{itemize}
    \item \textbf{AUC \& RMSE:} Adopting recognized metrics in the field of knowledge tracing. AUC (Area Under Curve) measures the model's ability to distinguish between "mastery" and "non-mastery," while RMSE (Root Mean Square Error) quantifies the deviation between predicted probabilities and actual response results.
    \item \textbf{Mistake Precision\citep{piech2015deep}:} Based on established standards, this evaluates not only whether the agent answers incorrectly but more importantly whether it commits "real and representative errors." This metric calculates the consistency between agent-generated incorrect options and typical student Misconceptions.
\end{itemize}

\subsubsection{Cognitive Dynamics:}
\begin{itemize}
    \item \textbf{Learning Curve Fitting ($R^2_{LC}$):} Based on relevant research\citep{newell1981mechanisms}, real human learning follows the Power Law of Practice. We calculate the goodness of fit $R^2$ between the agent's performance improvement trajectory and the power function. An $R^2 > 0.9$ typically represents extremely high psychological validity of the simulation data.
    \item \textbf{Behavior-Cognition Alignment (Alignment):} Evaluates the logical self-consistency between the agent's generated internal state (e.g., confidence/confusion) and its external performance (correct/incorrect).
\end{itemize}

\section{Evaluation and Discussion}

Table \ref{tab:main_results} presents the comprehensive performance of each model on CogMath-948. Experimental results reveal significant differences in fidelity across different modeling paradigms.

\begin{table}[ht]
    \centering
    \caption{Comparison of Ability Simulation and Cognitive Dynamics Metrics for Different Models on CogMath-948}
    \label{tab:main_results}
    \resizebox{\columnwidth}{!}{
    \begin{tabular}{lccccc}
        \toprule
        \textbf{Model} & \multicolumn{3}{c}{\textbf{Task Performance}} & \multicolumn{2}{c}{\textbf{Cognitive Dynamics}} \\
        \cmidrule(lr){2-4} \cmidrule(lr){5-6}
         & \textbf{AUC} $\uparrow$ & \textbf{RMSE} $\downarrow$ & \textbf{Mistake Prec.} $\uparrow$ & \textbf{$R^2_{LC}$} $\uparrow$ & \textbf{Align} $\uparrow$ \\
        \midrule
        \textit{Static Agents} & & & & & \\
        Llama 3 (\text{70B}) & 0.65 & 0.48 & 42.1\% & 0.45 & 0.72 \\
        GPT-4o & 0.71 & 0.43 & 55.4\% & 0.52 & 0.78 \\
        Gemini 3 Pro Think & 0.73 & 0.41 & 58.2\% & 0.61 & 0.82 \\
        \midrule
        \textit{KT-augmented} & & & & & \\
        PEERS (BKT+LLM) & \textbf{0.82} & \textbf{0.35} & 64.5\% & 0.78 & 0.75 \\
        \midrule
        \textit{Ours} & & & & & \\
        \textbf{CogEvolution} & 0.80 & 0.37 & \textbf{76.8\%} & \textbf{0.92} & \textbf{0.91} \\
        \bottomrule
    \end{tabular}
    }
\end{table}

\subsection{Analysis of Task Performance Accuracy (RQ1)}
In terms of mastery perception precision (AUC/RMSE), CogEvolution performs comparably to the knowledge tracing-based PEERS model (AUC: 0.80 vs 0.82), and both significantly outperform static persona models. This indicates that after introducing dynamic state update mechanisms, the agent can accurately capture fluctuations in student ability, similar to KT models. More importantly, CogEvolution achieves a significant advantage in \textbf{Mistake Precision} (76.8\%). Traditional Static Agents (such as GPT-4o) tend to generate "hallucinatory errors" or "logical leap errors," while PEERS generates error content that still relies on the base model despite predicting error probabilities. In contrast, through ICAP perception and memory retrieval, CogEvolution can reproduce typical Misconceptions consistent with real student cognitive gaps, demonstrating high fidelity in micro-level behavioral simulation.

\subsection{Learning Curve Fitting and Dynamics Analysis (RQ2)}

To quantitatively answer RQ2—whether the learning trajectories generated by CogEvolution conform to the fundamental laws of human cognition—we conducted a rigorous \textbf{Learning Curve Fitting} analysis based on the \textit{Power Law of Practice}.

\subsubsection{Metric Calculation Protocol}
In educational psychology, the learning rate of human students typically follows a non-linear decay pattern, where performance improves rapidly during early practice stages and gradually plateaus. We formalize this trajectory using the power function model:
\begin{equation}
    E(n) = A \cdot n^{-\alpha} + \epsilon
\end{equation}
where $E(n)$ represents the error rate at the $n$-th practice opportunity, $A$ is the initial error rate, and $\alpha$ denotes the learning rate decay parameter.
To evaluate the fidelity of the agents, we fit the error rate sequences generated by each model to this power function. The evaluation metric, \textbf{Learning Curve Fitting Degree ($R^2_{LC}$)}, is defined as the coefficient of determination ($R^2$) between the agent's simulated trajectory and the ground truth human trajectory from the CogMath-948 dataset:
\begin{equation}
    R^2_{LC} = 1 - \frac{\sum_{n} (y_{human}^{(n)} - y_{agent}^{(n)})^2}{\sum_{n} (y_{human}^{(n)} - \bar{y}_{human})^2}
\end{equation}
A higher $R^2_{LC}$ indicates that the agent's learning dynamics more closely mirror the "fast-then-slow" nature of human skill acquisition.

\subsubsection{Visualization and Result Discussion}
\begin{figure}[ht]
    \centering
    \includegraphics[width=0.9\linewidth]{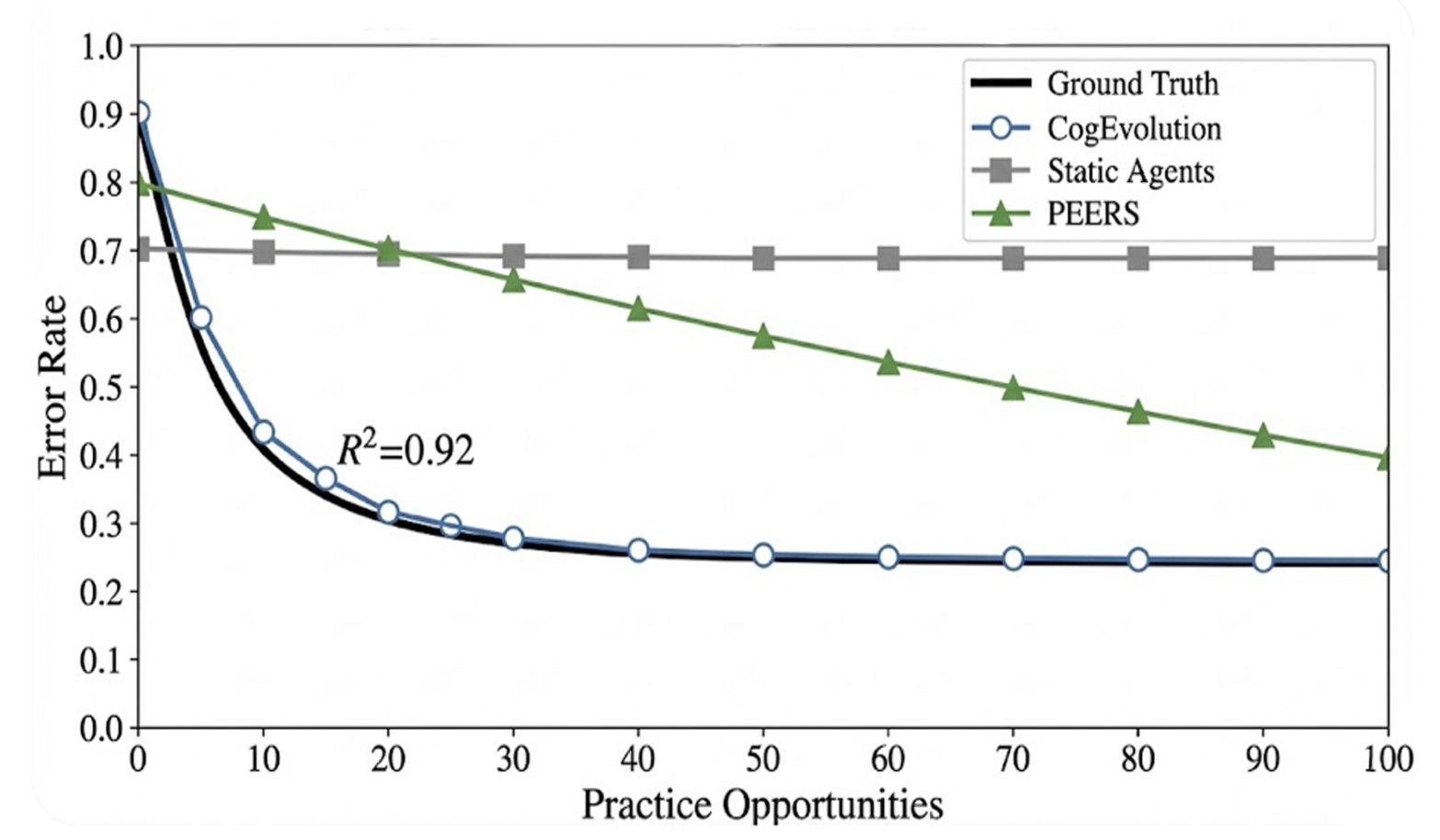} 
    \caption{Comparison of Learning Curves on CogMath-948. The black solid line represents the ground truth student performance, which strictly follows the Power Law of Practice. Specifically, the error rate of exercises related to the same knowledge point decreases as the number of practice opportunities increases until it approaches a convergence value based on the difficulty of the questions. The closer the learning curve simulated by the agent is to the actual situation, the more accurately it reflects the learner's knowledge level. This value is the computational indicator of learning curve fitting $R^2$. }
    \label{fig:learning_curve}
\end{figure}

Figure \ref{fig:learning_curve} visualizes the error rate trajectories of different agents over 100 practice opportunities. The visualization provides compelling evidence for the cognitive validity of CogEvolution.

As observed in Figure \ref{fig:learning_curve}, the \textbf{Static Agents} (Grey Square) exhibit a nearly flat trajectory ($R^2_{LC}=0.45$). This confirms that without a state update mechanism, standard LLMs rely solely on pre-trained knowledge, failing to manifest any "learning" effect from practice. 
The \textbf{PEERS} model (Green Triangle) shows a downward trend ($R^2_{LC}=0.78$), but its trajectory is overly linear. This suggests that while BKT-based probability transitions can simulate performance improvement, they struggle to capture the \textit{diminishing returns} characteristic of cognitive saturation.

In sharp contrast, \textbf{CogEvolution} (Blue Circle) produces a curve that tightly aligns with the Ground Truth ($R^2_{LC}=0.92$). The trajectory displays a steep drop in the first 20 steps, followed by a gradual stabilization. This high-fidelity simulation is attributed to our evolutionary update mechanism: early "Cognitive Spikes" (caused by Mutation) lead to rapid schema restructuring (Assimilation), while later updates become more refined (Fine-tuning), perfectly replicating the transition from "novice" to "proficient" as predicted by the Power Law.

\subsection{Ablation Study (RQ3)}
To verify the contribution of each module, we conducted ablation experiments (Table \ref{tab:ablation}).
\begin{itemize}
    \item \textbf{w/o ICAP:} Removing cognitive depth perception renders the agent unable to distinguish between shallow and deep learning, causing $R^2_{LC}$ to drop sharply to 0.58, demonstrating the validity of cognitive features in simulating agent cognitive psychological validity.
    \item \textbf{w/o Meta-Ret:} Removing structured retrieval leads to a 12.3\% decrease in Mistake Precision, indicating that knowledge assimilation is crucial for reproducing specific cognitive misconceptions.
    \item \textbf{w/o Evo-Update:} Removing the evolutionary update mechanism results in significantly reduced behavioral mistake fidelity and cognitive alignment (Align), causing the agent's simulation capability to fall back to static persona modeling.
\end{itemize}

\begin{table}[ht]
    \centering
    \caption{Ablation Study Results for Core CogEvolution Modules}
    \label{tab:ablation}
    \resizebox{0.9\columnwidth}{!}{
    \begin{tabular}{lccc}
        \toprule
        \textbf{Variant} & \textbf{Mistake Prec.} & \textbf{$R^2_{LC}$} & \textbf{Align} \\
        \midrule
        \textbf{CogEvolution (Full)} & \textbf{76.8\%} & \textbf{0.92} & \textbf{0.91} \\
        w/o ICAP & 68.5\% ($\downarrow$ 8.3) & 0.58 ($\downarrow$ 0.34) & 0.73 \\
        w/o Meta-Ret & 64.5\% ($\downarrow$ 12.3) & 0.85 ($\downarrow$ 0.07) & 0.88 \\
        w/o Evo-Update & 55.1\% ($\downarrow$ 21.7) & 0.51 ($\downarrow$ 0.41) & 0.76 \\
        \bottomrule
    \end{tabular}
    }
\end{table}

\section{Conclusion}
Addressing the pervasive issues of missing cognitive ability simulation and dynamic fluidity in generative educational agent research, this paper proposes the CogEvolution agent, which integrates an ICAP cognitive depth perceptron, structured memory retrieval, and a state update mechanism based on evolutionary strategies. Extensive experimental and simulation evaluations demonstrate that CogEvolution successfully realizes a paradigm shift from static behavioral simulation to dynamic cognitive evolution, marking a solid step toward constructing digital learners with "human-like cognition" and dynamic evolution.

\section*{Acknowledgments}
This research was supported by the General Program of the National Natural Science Foundation of China (No.~62377024), titled \textit{Discovering Interpretable Cognitive Evolution Paths Based on Causal Reasoning}, running from January~1,~2024 to December~31,~2027.

\nocite{*}

\printbibliography

\end{document}